\documentclass[conference]{IEEEtran}
\IEEEoverridecommandlockouts
\usepackage{cite}
\usepackage{url}
\usepackage{amsmath,amssymb,amsfonts}
\usepackage{algorithmic}
\usepackage{graphicx}
\usepackage{textcomp}
\usepackage{xcolor}
\usepackage{tikz}
\usepackage{pbox}
\usepackage{booktabs}
\usepackage{subcaption}
\def\BibTeX{{\rm B\kern-.05em{\sc i\kern-.025em b}\kern-.08em
    T\kern-.1667em\lower.7ex\hbox{E}\kern-.125emX}}

\begin{document}

\title{\textsc{GraphGen-Redux}: a Fast and Lightweight Recurrent Model for labeled Graph Generation
\thanks{Supported by the H2020 project TAILOR (n. 952215)}
}

\author{\IEEEauthorblockN{Davide Bacciu}
\IEEEauthorblockA{\textit{Dept. of Computer Science} \\
\textit{University of Pisa}\\
Largo Bruno Pontecorvo, 3 \\
davide.bacciu@unipi.it}
\and
\IEEEauthorblockN{Marco Podda}
\IEEEauthorblockA{\textit{Dept. of Computer Science} \\
\textit{University of Pisa}\\
Largo Bruno Pontecorvo, 3 \\
marco.podda@di.unipi.it}
}

\maketitle

\begin{abstract}
The problem of labeled graph generation is gaining attention in the Deep Learning community. The task is challenging due to the sparse and discrete nature of graph spaces. Several approaches have been proposed in the literature, most of which require to transform the graphs into sequences that encode their structure and labels and to learn the distribution of such sequences through an auto-regressive generative model. Among this family of approaches, we focus on the \textsc{GraphGen} model. The preprocessing phase of \textsc{GraphGen} transforms graphs into unique edge sequences called Depth-First Search (DFS) codes, such that two isomorphic graphs are assigned the same DFS code. Each element of a DFS code is associated with a graph edge: specifically, it is a quintuple comprising one node identifier for each of the two endpoints, their node labels, and the edge label. \textsc{GraphGen} learns to generate such sequences auto-regressively and models the probability of each component of the quintuple independently. While effective, the independence assumption made by the model is too loose to capture the complex label dependencies of real-world graphs precisely. By introducing a novel graph preprocessing approach, we are able to process the labeling information of both nodes and edges jointly. The corresponding model, which we term \textsc{GraphGen-Redux}, improves upon the generative performances of \textsc{GraphGen} in a wide range of datasets of chemical and social graphs. In addition, it uses approximately 78\% fewer parameters than the vanilla variant and requires 50\% fewer epochs of training on average.
\end{abstract}


\section{Introduction}
The task of labeled graph generation consists of learning the underlying probability distribution of a set of training labeled graphs, or a procedure to draw samples from it. This problem has several applications in real-world domains, for example in Cheminformatics \cite{jin2019multimodalmoltranslation,bradshaw2019moleculechef,podda2020aistats} and Network Science \cite{grover2019graphite,wang2018graphgan}, and has recently attracted a lot of attention in the Deep Learning community \cite{guo2020systematicreviewgenerativegraphs}, in conjunction with the development of Deep Generative Models (DGMs) able to learn inference and sampling from arbitrary data distributions. With respect to standard data domains where DGMs are applied such as images, sound, and text, generating graphs poses a set of unique challenges due to the complexity of graph spaces. In fact, graph spaces are (\textit{i}) \textit{sparse}, in the sense that only a small portion of them contains actually useful graphs; (\textit{ii}) \textit{discrete}, meaning that small moves in graph space (such as the deletion of a node from a graph) can have a dramatic impact on the outcome (e.g., the disconnection of a component); and (\textit{iii}) \textit{combinatorial}, which implies that brute-force enumeration of a graph space is usually intractable, especially in cases of variable-sized graphs with labeled nodes and edges \cite{bacciu2020dgn}. Thus, DGMs of graphs must cope with this intrinsic complexity in order to be successful in the task. 

Currently, one of the most effective models for labeled graph generation is \textsc{GraphGen} \cite{goyal2020graphgen}, a scalable model for domain-agnostic generation of both unlabeled and labeled graphs. The idea behind \textsc{GraphGen} is to combine Deep Learning with graph canonization. More precisely, the \textsc{GraphGen} framework first transforms graphs into unique sequences with a graph canonization algorithm. Each sequence element is associated to a graph edge, and it is a quintuple containing two timestamps identifying the order in which the endpoint nodes were discovered in a DFS visit, their labels, and the label of the edge connecting them. The distribution of these sequences is learned with a Long Short-Term Memory (LSTM) network \cite{hochreiter1997lstm}. In the original paper, \textsc{GraphGen} is shown to outperform other domain-agnostic baselines by a sensible margin on a large number of datasets and metrics. A simple observation on the process modeled by \textsc{GraphGen} reveals that all the components of the sequence elements (the quintuples) are generated independently. We argue that this assumption does not properly reflect the true distribution of the node and edge labels of real-world graphs, and develop a variant where the labels information is modeled jointly instead. We evaluate this variant thoroughly and fairly to vanilla \textsc{GraphGen}. The results show empirically than the proposed model, which we term \textsc{GraphGen-Redux}, generates more realistically-looking graphs coming from diverse training distribution. Moreover, it does so using on average 78\% less parameters and half the number of epochs required to train the vanilla version. 

\section{Background}
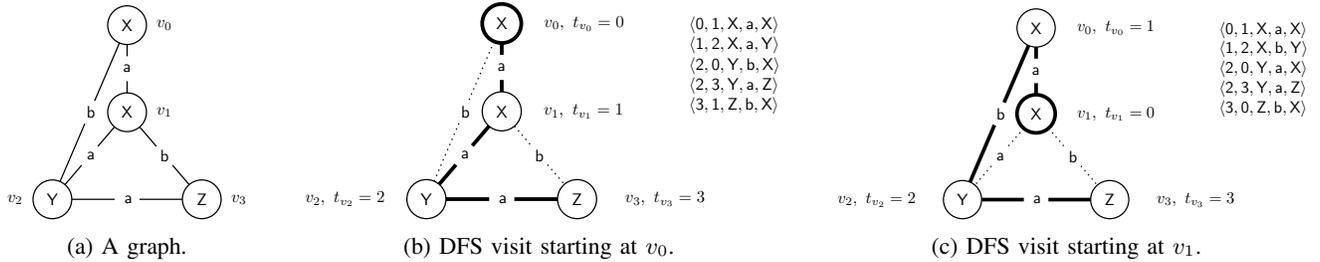
\begin{figure*}[t]
    \begin{subfigure}[b]{.21\linewidth}
        \centering
        \resizebox{.94\textwidth}{!}{\tikzset{every picture/.style={line width=0.75pt}} 

\begin{tikzpicture}[x=0.75pt,y=0.75pt,yscale=-1,xscale=1]

\draw    (270, 310) circle [x radius= 15.56, y radius= 15.56]   ;
\draw (270,310) node   [align=left] {\begin{minipage}[lt]{13.600000000000001pt}\setlength\topsep{0pt}
\begin{center}
$\displaystyle \mathsf{X}$
\end{center}

\end{minipage}};
\draw    (270, 380) circle [x radius= 15.56, y radius= 15.56]   ;
\draw (270,380) node   [align=left] {\begin{minipage}[lt]{13.600000000000001pt}\setlength\topsep{0pt}
\begin{center}
$\displaystyle \mathsf{X}$
\end{center}

\end{minipage}};
\draw    (210, 450) circle [x radius= 15.56, y radius= 15.56]   ;
\draw (210,450) node   [align=left] {\begin{minipage}[lt]{13.600000000000001pt}\setlength\topsep{0pt}
\begin{center}
$\displaystyle \mathsf{Y}$
\end{center}

\end{minipage}};
\draw    (330, 450) circle [x radius= 15.56, y radius= 15.56]   ;
\draw (330,450) node   [align=left] {\begin{minipage}[lt]{13.600000000000001pt}\setlength\topsep{0pt}
\begin{center}
$\displaystyle \mathsf{Z}$
\end{center}

\end{minipage}};
\draw (300,415) node   [align=left] {\begin{minipage}[lt]{8.67pt}\setlength\topsep{0pt}
\begin{center}
$\displaystyle \mathsf{b}$
\end{center}

\end{minipage}};
\draw (241,415) node   [align=left] {\begin{minipage}[lt]{8.67pt}\setlength\topsep{0pt}
\begin{center}
$\displaystyle \mathsf{a}$
\end{center}

\end{minipage}};
\draw (300,310) node   [align=left] {\begin{minipage}[lt]{13.600000000000001pt}\setlength\topsep{0pt}
\begin{center}
$\displaystyle v_{0}$
\end{center}

\end{minipage}};
\draw (300,380) node   [align=left] {\begin{minipage}[lt]{13.600000000000001pt}\setlength\topsep{0pt}
\begin{center}
$\displaystyle v_{1}$
\end{center}

\end{minipage}};
\draw (360,450) node   [align=left] {\begin{minipage}[lt]{13.600000000000001pt}\setlength\topsep{0pt}
\begin{center}
$\displaystyle v_{3}$
\end{center}

\end{minipage}};
\draw (180,450) node   [align=left] {\begin{minipage}[lt]{13.600000000000001pt}\setlength\topsep{0pt}
\begin{center}
$\displaystyle v_{2}$
\end{center}

\end{minipage}};
\draw (270,345) node   [align=left] {\begin{minipage}[lt]{8.67pt}\setlength\topsep{0pt}
\begin{center}
$\displaystyle \mathsf{a}$
\end{center}

\end{minipage}};
\draw (241,380) node   [align=left] {\begin{minipage}[lt]{8.67pt}\setlength\topsep{0pt}
\begin{center}
$\displaystyle \mathsf{b}$
\end{center}

\end{minipage}};
\draw (271,450) node   [align=left] {\begin{minipage}[lt]{8.67pt}\setlength\topsep{0pt}
\begin{center}
$\displaystyle \mathsf{a}$
\end{center}

\end{minipage}};
\draw    (270,325.56) -- (270,335)(270,355) -- (270,364.44) ;
\draw    (259.88,391.81) -- (246.51,407.41)(233.49,422.59) -- (220.12,438.19) ;
\draw    (280.12,391.81) -- (293.49,407.41)(306.51,422.59) -- (319.88,438.19) ;
\draw    (216.13,435.7) -- (236.06,389.19)(243.94,370.81) -- (263.87,324.3) ;
\draw    (225.56,450) -- (260,450)(280,450) -- (314.44,450) ;

\end{tikzpicture}}
        \caption{A graph.}
        \label{fig:dfs1}
    \end{subfigure}
    \begin{subfigure}[b]{0.38\linewidth}
        \centering
        \resizebox{.99\textwidth}{!}{\tikzset{every picture/.style={line width=0.75pt}} 

\begin{tikzpicture}[x=0.75pt,y=0.75pt,yscale=-1,xscale=1]

\draw  [line width=2.25]   (360, 310) circle [x radius= 15.56, y radius= 15.56]   ;
\draw (360,310) node   [align=left] {\begin{minipage}[lt]{13.600000000000001pt}\setlength\topsep{0pt}
\begin{center}
$\displaystyle \mathsf{X}$
\end{center}

\end{minipage}};
\draw    (360, 380) circle [x radius= 15.56, y radius= 15.56]   ;
\draw (360,380) node   [align=left] {\begin{minipage}[lt]{13.600000000000001pt}\setlength\topsep{0pt}
\begin{center}
$\displaystyle \mathsf{X}$
\end{center}

\end{minipage}};
\draw    (300, 450) circle [x radius= 15.56, y radius= 15.56]   ;
\draw (300,450) node   [align=left] {\begin{minipage}[lt]{13.600000000000001pt}\setlength\topsep{0pt}
\begin{center}
$\displaystyle \mathsf{Y}$
\end{center}

\end{minipage}};
\draw    (420, 450) circle [x radius= 15.56, y radius= 15.56]   ;
\draw (420,450) node   [align=left] {\begin{minipage}[lt]{13.600000000000001pt}\setlength\topsep{0pt}
\begin{center}
$\displaystyle \mathsf{Z}$
\end{center}

\end{minipage}};
\draw (390,415) node   [align=left] {\begin{minipage}[lt]{8.67pt}\setlength\topsep{0pt}
\begin{center}
$\displaystyle \mathsf{b}$
\end{center}

\end{minipage}};
\draw (331,415) node   [align=left] {\begin{minipage}[lt]{8.67pt}\setlength\topsep{0pt}
\begin{center}
$\displaystyle \mathsf{a}$
\end{center}

\end{minipage}};
\draw (425.5,310) node   [align=left] {\begin{minipage}[lt]{60.52pt}\setlength\topsep{0pt}
\begin{center}
$\displaystyle v_{0} ,\ t_{v_{0}} =0$
\end{center}

\end{minipage}};
\draw (425,380) node   [align=left] {\begin{minipage}[lt]{61.2pt}\setlength\topsep{0pt}
\begin{center}
$\displaystyle v_{1} ,\ t_{v_{1}} =1$
\end{center}

\end{minipage}};
\draw (490,450) node   [align=left] {\begin{minipage}[lt]{68pt}\setlength\topsep{0pt}
\begin{center}
$\displaystyle v_{3} ,\ t_{v_{3}} =3$
\end{center}

\end{minipage}};
\draw (235,450) node   [align=left] {\begin{minipage}[lt]{61.2pt}\setlength\topsep{0pt}
\begin{center}
$\displaystyle v_{2} ,\ t_{v_{2}} =2$
\end{center}

\end{minipage}};
\draw (360,345) node   [align=left] {\begin{minipage}[lt]{8.67pt}\setlength\topsep{0pt}
\begin{center}
$\displaystyle \mathsf{a}$
\end{center}

\end{minipage}};
\draw (331,380) node   [align=left] {\begin{minipage}[lt]{8.67pt}\setlength\topsep{0pt}
\begin{center}
$\displaystyle \mathsf{b}$
\end{center}

\end{minipage}};
\draw (361,450) node   [align=left] {\begin{minipage}[lt]{8.67pt}\setlength\topsep{0pt}
\begin{center}
$\displaystyle \mathsf{a}$
\end{center}

\end{minipage}};
\draw (501,302) node [anchor=north west][inner sep=0.75pt]   [align=left] {$\displaystyle 
\begin{array}{l}
\mathsf{\langle 0,1,X,a,X\rangle }\\
\mathsf{\langle 1,2,X,a,Y\rangle }\\
\mathsf{\langle 2,0,Y,b,X\rangle }\\
\mathsf{\langle 2,3,Y,a,Z\rangle }\\
\mathsf{\langle 3,1,Z,b,X\rangle }
\end{array}$};
\draw [line width=2.25]    (360,325.56) -- (360,335)(360,355) -- (360,364.44) ;
\draw [line width=2.25]    (349.88,391.81) -- (336.51,407.41)(323.49,422.59) -- (310.12,438.19) ;
\draw [line width=0.75]  [dash pattern={on 0.84pt off 2.51pt}]  (370.12,391.81) -- (383.49,407.41)(396.51,422.59) -- (409.88,438.19) ;
\draw  [dash pattern={on 0.84pt off 2.51pt}]  (306.13,435.7) -- (326.06,389.19)(333.94,370.81) -- (353.87,324.3) ;
\draw [line width=2.25]    (404.44,450) -- (370,450)(350,450) -- (315.56,450) ;

\end{tikzpicture}}
        \caption{DFS visit starting at $v_0$.}
        \label{fig:dfs2}
    \end{subfigure}
    \begin{subfigure}[b]{0.38\linewidth}
        \centering
        \resizebox{.99\textwidth}{!}{\tikzset{every picture/.style={line width=0.75pt}} 

\begin{tikzpicture}[x=0.75pt,y=0.75pt,yscale=-1,xscale=1]

\draw  [line width=0.75]   (190, 240) circle [x radius= 15.56, y radius= 15.56]   ;
\draw (190,240) node   [align=left] {\begin{minipage}[lt]{13.600000000000001pt}\setlength\topsep{0pt}
\begin{center}
$\displaystyle \mathsf{X}$
\end{center}

\end{minipage}};
\draw  [line width=2.25]   (190, 310) circle [x radius= 15.56, y radius= 15.56]   ;
\draw (190,310) node   [align=left] {\begin{minipage}[lt]{13.600000000000001pt}\setlength\topsep{0pt}
\begin{center}
$\displaystyle \mathsf{X}$
\end{center}

\end{minipage}};
\draw    (130, 380) circle [x radius= 15.56, y radius= 15.56]   ;
\draw (130,380) node   [align=left] {\begin{minipage}[lt]{13.600000000000001pt}\setlength\topsep{0pt}
\begin{center}
$\displaystyle \mathsf{Y}$
\end{center}

\end{minipage}};
\draw    (250, 380) circle [x radius= 15.56, y radius= 15.56]   ;
\draw (250,380) node   [align=left] {\begin{minipage}[lt]{13.600000000000001pt}\setlength\topsep{0pt}
\begin{center}
$\displaystyle \mathsf{Z}$
\end{center}

\end{minipage}};
\draw (220,345) node   [align=left] {\begin{minipage}[lt]{8.67pt}\setlength\topsep{0pt}
\begin{center}
$\displaystyle \mathsf{b}$
\end{center}

\end{minipage}};
\draw (161,345) node   [align=left] {\begin{minipage}[lt]{8.67pt}\setlength\topsep{0pt}
\begin{center}
$\displaystyle \mathsf{a}$
\end{center}

\end{minipage}};
\draw (255,240) node   [align=left] {\begin{minipage}[lt]{61.2pt}\setlength\topsep{0pt}
\begin{center}
$\displaystyle v_{0} ,\ t_{v_{0}} =1$
\end{center}

\end{minipage}};
\draw (190,275) node   [align=left] {\begin{minipage}[lt]{8.67pt}\setlength\topsep{0pt}
\begin{center}
$\displaystyle \mathsf{a}$
\end{center}

\end{minipage}};
\draw (255,310) node   [align=left] {\begin{minipage}[lt]{61.2pt}\setlength\topsep{0pt}
\begin{center}
$\displaystyle v_{1} ,\ t_{v_{1}} =0$
\end{center}

\end{minipage}};
\draw (320,380) node   [align=left] {\begin{minipage}[lt]{68pt}\setlength\topsep{0pt}
\begin{center}
$\displaystyle v_{3} ,\ t_{v_{3}} =3$
\end{center}

\end{minipage}};
\draw (60,380) node   [align=left] {\begin{minipage}[lt]{68pt}\setlength\topsep{0pt}
\begin{center}
$\displaystyle v_{2} ,\ t_{v_{2}} =2$
\end{center}

\end{minipage}};
\draw (161,310) node   [align=left] {\begin{minipage}[lt]{8.67pt}\setlength\topsep{0pt}
\begin{center}
$\displaystyle \mathsf{b}$
\end{center}

\end{minipage}};
\draw (191,380) node   [align=left] {\begin{minipage}[lt]{8.67pt}\setlength\topsep{0pt}
\begin{center}
$\displaystyle \mathsf{a}$
\end{center}

\end{minipage}};
\draw (331,232) node [anchor=north west][inner sep=0.75pt]   [align=left] {$\displaystyle  \begin{array}{{l}}
\mathsf{\langle 0,1,X,a,X\rangle }\\
\mathsf{\langle 1,2,X,b,Y\rangle }\\
\mathsf{\langle 2,0,Y,a,X\rangle }\\
\mathsf{\langle 2,3,Y,a,Z\rangle }\\
\mathsf{\langle 3,0,Z,b,X\rangle }
\end{array}$};
\draw [line width=2.25]    (190,255.56) -- (190,265)(190,285) -- (190,294.44) ;
\draw [line width=0.75]  [dash pattern={on 0.84pt off 2.51pt}]  (179.88,321.81) -- (166.51,337.41)(153.49,352.59) -- (140.12,368.19) ;
\draw [line width=0.75]  [dash pattern={on 0.84pt off 2.51pt}]  (200.12,321.81) -- (213.49,337.41)(226.51,352.59) -- (239.88,368.19) ;
\draw [line width=2.25]    (136.13,365.7) -- (156.06,319.19)(163.94,300.81) -- (183.87,254.3) ;
\draw [line width=2.25]    (145.56,380) -- (180,380)(200,380) -- (234.44,380) ;

\end{tikzpicture}}
        \caption{DFS visit starting at $v_1$.}
        \label{fig:dfs3}
    \end{subfigure}
    \caption{An example graph and two of its possible DFS codes (displayed on the right). The starting nodes of the DFS visit and the corresponding forward edges are highlighted with thicker strokes. The DFS code in (b) is also the m-DFS code of the graph depicted in (a).}
    \label{fig:dfs-code}
\end{figure*}
\subsection{Notation and Problem}\label{sec:notation}
In this section, we introduce the necessary notation and describe the problem of sequential graph generation. Let $G = \langle V_G, E_G\rangle$ be a graph with $n$ nodes $V_G = \{v_1, \ldots, v_n\}$ and edges $E_G = \{(v_i,v_j) \mid v_i, v_j \in V \}$. In this work, we assume undirected graphs with no self-loops and discrete node and edge labels. In particular, we posit the existence of a function $L_{node}: V \to \mathcal{V}$ that maps nodes to a set $\mathcal{V}$ of $|\mathcal{V}|$ possible node labels; analogously, we posit a function $L_{edge}: E \to \mathcal{E}$ that maps edges to a set $\mathcal{E}$ of $|\mathcal{E}|$ possible edge labels. Given two graphs $G$ and $G'$, we say that they are isomorphic if there exists a bijection $\phi: V_G \to V_{G'}$ such that $(u,v) \in E_G$ if and only if $(\phi(u), \phi(v)) \in E_{G'}$. Since we assume all graphs are labeled, we also require $L_{node}(v) = L_{node}(\phi(v)), \forall v \in V_G$ and $L_{edge}(e) = L_{edge}(e')$ with $e' = (\phi(v_i),\phi(v_j)), \forall e = (v_i,v_j) \in E_G$. We define the \textit{canonical label} of a graph as a string of text such that, given two graphs $G$ and $G'$, they are assigned the same canonical label if and only if they are isomorphic. We remark that finding a canonical labeling has the same asymptotic cost as to determine whether two graphs are isomorphic \cite{babai1983canononicalgraph}. 

The problem of graph generation is ascribable to learning a model $p_{model}(G)$, which approximates some true unknown graph distribution $p(G)$ using a dataset $\mathcal{G} = \{G_i\}_{i=1}^N$ of graphs drawn i.i.d. from it. This problem is hard in principle, since a graph is invariant to the permutation of its nodes. One approach to circumvent this complexity is to relax the permutation invariant requirement and model the graph generation as a sequential process. Let $\mathcal{F}$ be a mapping that converts graphs $G$ into sequences, and let $S$ be the sequence resulting from applying $\mathcal{F}$ to $G$. Sequential graph generation replaces $p(G)$ with $p(S)$, and learns an approximation $p_{model}(S)$ using a dataset $\mathcal{S} = \{S_i = \mathcal{F}(G_i) \mid G_i \in \mathcal{G}\}$.

\subsection{Minimum DFS-codes}\label{sec:min-dfscodes}
The desirable approach to sequential generation is to convert a graph into a sequence that is also its canonical form. Specifically, if the mapping $\mathcal{F}$ produces a canonical form, then it is invertible, and one can recover the original graph from the sequence without errors. Several algorithms have been devised to produce canonical forms out of graphs \cite{mckay2014nauty}. Here, we focus on an algorithm which does so through so-called \textit{DFS codes}. To obtain a DFS code of a graph $G$, the first step consists of performing a DFS visit of the graph starting with one node at random. A \textit{timestamp} corresponding to the order of visit is assigned to each node in the graph. Once timestamps are assigned, the generic edge $e=(u,v) \in E_G$ can be rewritten as a quintuple $\langle t_u, t_v, L_{node}(u), L_{edge}(e), L_{node}(v)\rangle$, where $t_u, t_v \in \mathbb{N}$ denote the timestamps assigned to the endpoints $u$ and $v$, respectively. The DFS visit partitions the set of edges of the graph in one set of \textit{forward} edges, traversed during the visit, and one set of \textit{backward} edges, not traversed during the visit. Note that if $(u,v)$ is a forward edge, we have $t_u < t_v$. Similarly, for backward edges, it holds $t_u > t_v$. These two partitions can be used to enforce a total ordering among the graph edges according to the following rules:
\begin{itemize}
    \item if $(u,v)$ is a backward edge, it is placed \textit{before} any forward edge whose first endpoint is $u$;
    \item if $(u,v)$ is a backward edge, it is placed \textit{after} any forward edge whose second endpoint is $u$;
    \item if $(u, v)$ and $(u, w)$ are two backward edges and $t_v < t_w$, $(u,v)$ is placed before $(u,w)$.
\end{itemize}
Any sequence of quintuples where $t_u$ and $t_v$ are ordered according to these rules is a valid DFS code; it follows that the number of DFS codes of a graph is equal to the number of possible DFS visits. In Figg.~\ref{fig:dfs2}-\ref{fig:dfs3}, we show two possible DFS codes for the graph of Fig.~\ref{fig:dfs1}, derived from a DFS starting at node $v_0$ and $v_1$, respectively. The \textit{lexicographically} smaller code among all the DFS codes is called the \textit{minimum} DFS (m-DFS) code of the graph. The DFS code of Fig.~\ref{fig:dfs2} is a m-DFS code. As it turns out, any graph can be reduced in canonical form through its m-DFS code \cite{yan2002gspan}. Clearly, the computational cost of finding the m-DFS code is $O(n!)$ in the worst case, which corresponds to evaluating all possible node permutations to find the m-DFS code. However, in practical cases (and more so in the case of labeled graphs) the search space is usually much smaller, and the computational cost of finding the m-DFS code is sustainable \cite{goyal2020graphgen}. For example, in Fig.~\ref{fig:dfs-code}, only two derivation steps are required to decide that the DFS code of Fig.~\ref{fig:dfs2} is smaller than that of Fig.~\ref{fig:dfs2}, since the second quintuple of the first DFS code $\langle 1,2,X,a,Y\rangle$ is lexicographically smaller than the second quintuple of the second DFS code $\langle 1,2,X,b,Y\rangle$. If the graph is unlabeled, graph invariants such as node degrees can be used as labels to prune the search space and speed up the process. Thus, in this work, $\mathcal{F}$ is the algorithm that constructs a given m-DFS code from an input graph.

\subsection{Domain-agnostic graph generation}
Among the approaches to approximate graph distributions from data with Deep Learning methodologies, we distinguish two broad families: the first one concerns modelling the entries of the graph adjacency matrix as independent Bernoulli random variables, so that a graph can be generated by sampling them in parallel \cite{decao2018molgan,simonovsky2018graphvae}; the second one concerns modelling graph generation as a sequential process that adds nodes, edges, or subgraphs to an initially empty graph \cite{johnson2017learngraphtrans,li2018learningdeepgmg,you2018graphrnn}. In this work, we focus on the latter approach, which grants a superior expressiveness (due to the possibility of modelling complex node and edge dependencies more effectively), but is computationally more demanding, since training and sampling cannot be parallelized in general. Moreover, there is the problem of how to enforce a consistent ordering of the operations to the sequence elements across different graphs. Sequential DGMs of graphs can be further distinguished in node-based, edge-based and block-based. Focusing on domain-agnostic generators, which are of interest in this work, one of the first approaches is the Deep Generative Model of Graphs (DGMG) \cite{li2018learningdeepgmg}, which conceives the generative process of a graph as a sequence of decisions parameterized by deep networks. At each step, the model first decides whether to add a new node to an existing graph; if so, it then decides to which nodes it is connected to, and predicts the node and edge labels accordingly. The state of the graph being generated is maintained with a Graph Neural Network (GNN) \cite{scarselli2009gnn,micheli2009nn4g,bacciu2018contgraphmarkov}. The authors do not provide an exhaustive solution to the ordering problem, but they experiment with different ordering strategies, showing that they work empirically. GraphRNN \cite{you2018graphrnn} generates graphs using two networks: a node-based RNN responsible of updating the state of the graph generated so far, and an edge-based network for generating an adjacency vector to connect a new node to the existing graph. The original formulation was later extended in \cite{popova2019molecularrnn} to deal with labeled graphs. Here, the model is trained on every possible node ordering, which requires to train the model for a large number of epochs to converge. Among edge-based, domain-agnostic approaches, the model in \cite{bacciu2019edgegraphgenrnn,bacciu2019graphesann} generates graphs by splitting their edge sequence in two sub-sequences: one containing the first endpoints of the edges, and another containing the second endpoints. The two sequences are then learned by two RNNs in cascade. In this case, the authors adopt an ordering strategy derived from a BFS visit of the graphs, which works empirically despite not being theoretically well-founded. One other limitation of this model is that it can only generate unlabeled graphs. Finally, the Graph Recurrent Attention Network (GRAN) \cite{kawai2019scalablegran} generates and adds nodes to an existing graph in blocks, which results in an improved scalability. The model uses an attention-based GNN to connect the block to the existing graphs, mitigating (but not solving completely) the dependency on the node ordering with the permutation invariance typical of graph-based neural models. However, GRAN cannot handle node and edge labels.

\subsection{\textsc{GraphGen} and its limitations}
Our work is based on \textsc{GraphGen} \cite{goyal2020graphgen}, a domain-agnostic and scalable model for labeled sequential graph generation. \textsc{GraphGen} first preprocesses the graphs in a dataset and transforms them into sequences of edges through their m-DFS codes, as explained in Sec.~\ref{sec:min-dfscodes}. This effectively transforms the graph dataset into a dataset of sequences, whose distribution is learned using a RNN with LSTM cells. The advantage of \textsc{GraphGen} is that it uses a canonical labeling algorithm, which allows the model to be trained on a one-to-one relation between the graphs and the sequences, in contrast to the previous approaches. The authors show experimentally the success of the framework with a thorough evaluation against \textsc{GraphRNN} and \textsc{DGMG}. The main limitation of the \textsc{GraphGen} model stems from the way the quintuples constituting the m-DFS code are generated. The approach adopted in the paper is to model each quintuple element independently. However, independence is often unsuited to model real-world relationships between the node and edge labels. To illustrate this concept, consider a citation network where nodes represent authors and there is a link between two authors if they coauthored an article together. Suppose nodes are labeled with the field of expertise of the corresponding author, and that edges are labeled with the subject of the article. If a link between a biologist and a computer scientist is observed, the subject of the article becomes more likely to be about computational biology than, say, philosophy. As another example, in chemical graphs Fluorine atoms cannot physically form double bonds. Thus, whenever a Fluorine atom is encountered, the possibility that the Fluorine atom is connected to any other atom with a double bond is immediately ruled out: and in fact, the presence of a double bond attached to a Fluorine atom would result in a chemically invalid molecule. In all these practical cases, assuming independence between node and edge labels might be insufficient to generate realistic graphs, or even result in generating nonsensical graphs (in the latter case, from a chemical point of view). In the following sections, we propose a solution to practically address this limitation.

\section{Methods}
\subsection{Preprocessing}
\begin{figure*}[t]
    \centering
    \resizebox{.65\textwidth}{!}{\input{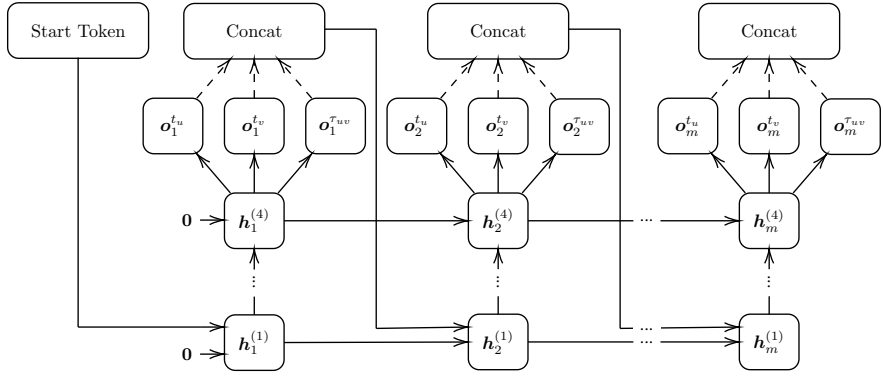}}
    \caption{A depiction of the generative process of the proposed model. Dashed arrows indicate a draws from the corresponding output distributions. The sample is one-hot encoded before concatenation. The linear embedding layer (which connects the input to the first LSTM layer) is not shown.}
    \label{fig:model}
\end{figure*}
In light of the previous observations, we modify the preprocessing step of \textsc{GraphGen} to enforce the required dependencies. Starting out with a graph dataset $\mathcal{G} = \{G_i\}_{i=1}^n$, we transform them into m-DFS codes, obtaining a sequence dataset $\{S_i = \mathcal{F}(G_i) \mid G \in \mathcal{G}\}$, where $S_i = (s_1^{(i)}, \ldots, s_{|E_G|}^{(i)})$ is the m-DFS code obtained as described in Sec.~\ref{sec:min-dfscodes}. Recall that each sequence element $s_i$ is a quintuple $\langle  t_u,t_v,L_{node}(u),L_{edge}(e),L_{node}(v)\rangle$ with $u, v \in V_G$ and $e = (u,v) \in E_G$. At this point, we the preprocessing phase is the same as the one proposed in the \textsc{GraphGen} paper. We choose to enforce a dependency between the label information contained in the quintuple algorithmically, by converting them to triplets of the form $s_i = \langle t_u, t_v, \tau_{uv} \rangle$ where $\tau_{uv} = L_{node}(u)L_{edge}(e)L_{node}(v)$. In words, we transform the disjoint information about the node labels and the edge label into a unique \textit{token} $\tau_{uv}$ by concatenating them together. Note that the property of the m-DFS code of being a canonical graph label is not lost with this simple modification, since the lexicographic order is maintained. Although this modification is simple, it has the profound implication of allowing training the model to learn the node and edge labels jointly. We call the m-DFS code transformed in triplet notation a \textit{reduced} m-DFS code. Note that, after the modification to the preprocessing, the number of possible node and edge labels increases, as now each combination of $L_{node}(u)$, $L_{edge}(e)$, and $L_{edge}(v)$ found in the dataset becomes a token label, with an upper bound of $|\mathcal{V}| \times |\mathcal{V}| \times |\mathcal{E}|$ possible combinations. However, we found out empirically that the number of actual combinations in the datasets on which we conducted the experiments is much smaller than the upper bound.

\subsection{Model}
As anticipated in Sec.~\ref{sec:notation}, our task is to learn $p(S)$, which in our case corresponds to the probability distribution of the reduced m-DFS codes, using a dataset $\mathcal{S}$. We model the probability of sampling a sequence $S$ (with length $|S|$) from the data distribution $p(S)$ with the following conditional:
\begin{align}
    p(S) = \prod_{i=1}^{|S|+1}p(s_i | s_1, \ldots, s_{i-1}) = \prod_{i=1}^{|S|+1}p(s_i | s_{<i}),
\end{align}
where we add +1 to the product to account for an EOS token that signals the end of the sequence. In practice, the learning problem is reduced to an autoregressive task: the objective is to learn the distribution of each triplet $s_i = (t_u, t_v, \tau_{uv})$, conditioned on the previous generative history $s_{<i}$. We assume that the two timestamps $t_u$ and $t_v$, and the token $\tau_{uv}$ are generated independently from each other. Thus, the conditional distribution factors out as follows:
\begin{align}
\begin{split}
    p(s_i|s_{<i}) &= p(\langle t_u, t_v, \tau_{uv}\rangle | s_{<i})\\
    &=p(t_u|s_{<i}) \times p(t_v|s_{<i}) \times p(\tau_{uv}|s_{<i}).
\end{split}
\end{align}
Note that, with respect to the \textsc{GraphGen} model, the term $p(\tau_{uv}|s_{<i})$ now specifies the \textit{joint} distribution of the node and edge labels associated to the sequence element. In contrast, the generation of the two endpoint labels and the edge label are independent in \textsc{GraphGen}. The propose to learn this distribution with the following Deep Learning model:
\begin{align}
\begin{split}
    \boldsymbol{h}_i &= \mathrm{RNN}(\boldsymbol{h}_{i-1}, \mathrm{Emb}(\boldsymbol{s}_{i-1}))\\
    \boldsymbol{o}_i^{t_u} &= \mathrm{MLP}^{t_u}(\boldsymbol{h}_i)\\
    \boldsymbol{o}_i^{t_v} &= \mathrm{MLP}^{t_v}(\boldsymbol{h}_i)\\
    \boldsymbol{o}_i^{\tau_{uv}} &= \mathrm{MLP}^{\tau_{uv}}(\boldsymbol{h}_i).
\end{split}
\end{align}
Above, Emb is a linear embedding layer, RNN is a recurrent network composed of four LSTM layers, and the three MLPs map the output of the last LSTM layer to the corresponding output distribution (of the two timestamps and the token) by applying a softmax function to their outputs. In contrast with the original \textsc{GraphGen} implementation, our model requires only three MLPs (instead of five). The model is initialized with $\boldsymbol{h}_0 = \boldsymbol{0}$, the zero vector, and with $\boldsymbol{s}_0 = \langle \mathtt{S} \rangle$, an SOS token. The three distributions given by the MLPs are finally concatenated together to construct the per-step output of the network:

\begin{align}
    \hat{\boldsymbol{s}}_i = \mathrm{Concat}(\boldsymbol{o}_i^{t_u}, \boldsymbol{o}_i^{t_v}, \boldsymbol{o}_i^{\tau_{uv}}).
\end{align}

The model is trained to minimize the Binary Cross-Entropy (BCE) loss between the network output at each step and the corresponding ground truth sequence element. More specifically:

\begin{align}
    \begin{split}
        \mathrm{BCE}(\hat{\boldsymbol{s}}_i, \boldsymbol{s}_i) = -\sum_{i=1}^{|S|+1} \boldsymbol{s}_i \log(\hat{\boldsymbol{s}}_i) + (1-\boldsymbol{s}_i) \log(1-\hat{\boldsymbol{s}}_i),
    \end{split}
\end{align}

where $\boldsymbol{s}_i$ is the vector obtained by concatenating the one-hot encoded vectors representing $t_u$, $t_v$, and $\tau_{uv}$ respectively. During training, we use teacher forcing \cite{williams1989teacherforcing} and feed the ground truth vector $\boldsymbol{s}_i$ instead of the network output $\hat{\boldsymbol{s}}_i$ as input to the next step of the recurrence.

\subsection{Generation}
The generative phase of the model starts by feeding $\boldsymbol{h}_0$ and the one-hot encoded SOS token $\boldsymbol{s}_0$ to the network, which computes the output distributions of the two timestamps and the token. These distributions are sampled, the sampled values are turned into one-hot encoded vectors, and concatenated together to form the input for the next step of the recurrence. The process is iterated until an EOS token $\langle \mathtt{E} \rangle$ is sampled from any of the three output distributions, at which point the generative process is interrupted. The result is a reduced m-DFS sequence, which is converted back into the corresponding graph. Fig.~\ref{fig:model} shows the generative process visually.

\section{Experiments}\label{sec:experiments}
The experimental section of this paper aims at demonstrating empirically the following claims about \textsc{GraphGen-Redux} with respect to the original \textsc{GraphGen} model:
\begin{itemize}
    \item it has superior generative performances;
    \item it is more lightweight, in terms of number of parameters needed to achieve the superior performance;
    \item it is faster, in terms of training time.
\end{itemize}
To support the first claim, we show that the model consistently outperforms the \textsc{GraphGen} baseline on eleven metrics which access local and global structural similarity, label accounting, and redundancy. \textsc{GraphGen-Redux} obtains these results with 78\% (on average) less than \textsc{GraphGen}, which substantiates the second claim. Moreover, we also show that training \textsc{GraphGen} with the same number of parameters as our model worsens its performances. Finally, to support the third claim, we show that \textsc{GraphGen-Redux} outperforms \textsc{GraphGen} with approximately half the number of epochs.

\subsection{Datasets}
We experiment on five different datasets, which were originally used to evaluate the \textsc{GraphGen} model. The datsets are described in the following.
\begin{itemize}
    \item \textsc{PubChem}: a merged collection of four different chemical datasets, where graphs represent molecules, taken from PubChem \cite{gindulyte2016pubchem}. It comprises 80255 graphs;
    \item \textsc{Lung}: a subset of \textsc{PubChem} dataset  comprising 24835 graphs;
    \item \textsc{Yeast}: a subset of \textsc{PubChem} dataset comprising 55533 graphs;
    \item \textsc{Enzymes}: a dataset of 575 graphs from the Brenda enzyme database \cite{schomburg2004enzymes}. In this dataset, we complement the labels of each node with the corresponding node degree to facilitate the algorithm for constructing the m-DFS code;
    \item \textsc{CiteSeer}: a dataset of graphs extracted from the CiteSeer citation network \cite{giles1998citeseer}. Since CiteSeer consists of one large graph, we used the following procedure to construct the dataset. First, we select a node at random with probability proportional to its degree. Then, we perform 150 random walks starting from that node with restart probability of 0.15. Any edge traversed during the walk is included in the sampled subgraph. The obtained dataset comprises 13348 graphs;
    \item \textsc{Cora}: a dataset of graphs extracted from the Cora citation network \cite{sen2008citeseercora}, constructed using the same procedure described for \textsc{CiteSeer}. The dataset comprises 18836 graphs.
\end{itemize}
Tab.~\ref{tab:data} summarizes the information about the dataset. In particular, the second column displays the maximum number of nodes (which corresponds to the output dimention of $\mathrm{MLP}^{t_u}$ and $\mathrm{MLP}^{t_v}$), and the number of possible token (the output dimension of $\mathrm{MLP}^{\tau_{uv}}$). Notice that the number of tokens is usually smaller than the actual number of combinations of possible node and edge labels. For example, for the \textsc{PubChem} dataset, we found only 80 possible tokens out of 576 possible combinations. 

For all these datasets, we reserve 10\% of the graphs as test set for evaluation purposes, and 10\% of the graphs remaining after having separated the test set for validation purposes. Importantly, we \textit{fix} the data splits to ensure reproducibility and fairness in the comparison\cite{errica2020fairevaluation}. For the \textsc{Cora} and \textsc{CiteSeer} datasets, we fix the random seed during their construction. 
\begin{table}[t]
    \begin{center}
    \begin{tabular}{l c c c }
    \toprule
    \parbox[h]{1cm}{\centering Dataset}& 
    \parbox[h]{1cm}{\centering Num.\\graphs}& 
    \parbox[h]{1cm}{\centering $t_u, t_v$\\dim.}& 
    \parbox[h]{1.5cm}{\centering $\tau_{uv}$ dim.\\Actual(Total)}\\
    \midrule
    \textsc{PubChem} & 80255 & 130 & 80(576) \\
    \textsc{Lung} & 24835 & 51 & 71(576) \\
    \textsc{Yeast} & 55533 & 51 & 71(576) \\
    \textsc{Enzymes} & 575 & 125 & 354(530) \\
    \textsc{CiteSeer} & 13348 & 90 & 36(49) \\
    \textsc{Cora} & 18836 & 103 & 48(64) \\
    \bottomrule
    \end{tabular}
    \caption{Dataset Statistics.} \label{tab:data}
    \end{center}
\end{table}
\begin{table*}[t]
    \scriptsize
    \begin{center}
    \begin{tabular}{l l c c c c c c c c c c c c c}
    \toprule
    \parbox[h]{0.8cm}{\centering Dataset}& 
    \parbox[h]{1.3cm}{\centering Model}&  
    \parbox[h]{0.7cm}{\centering Degree}& 
    \parbox[h]{0.6cm}{\centering Clust.}& 
    \parbox[h]{0.6cm}{\centering Orbit}& 
    \parbox[h]{0.7cm}{\centering Avg.\\Nodes}&
    \parbox[h]{0.6cm}{\centering Avg.\\Edges}&
    \parbox[h]{0.7cm}{\centering Node\\Labels}&
    \parbox[h]{0.7cm}{\centering Edge\\Labels}&
    \parbox[h]{0.8cm}{\centering Node\\ Labels+\\Degree}& 
    \parbox[h]{0.8cm}{\centering NSPDK}& 
    \parbox[h]{0.7cm}{\centering Novel}& 
    \parbox[h]{0.72cm}{\centering Unique}&
    \parbox[h]{0.7cm}{\centering Valid}&
    \parbox[h]{0.5cm}{\centering Rank}\\

    \midrule
                     & \textsc{GraphRNN}       & 0.241 & $\approx \boldsymbol{0}$ & 0.039 & 8.2/37.6 & 7.2/39.1 & 0.102 & 0.010 & 0.879 & 0.331 & 62.0\% & 52.0\% & - & 4\\
                     & \textsc{DGMG}           &  0.074 & 0.002 & 0.002 & 24.3/37 & 24.1/39.1 & 0.092 & 0.002 & 0.912 & 0.221 & 83.0\% & 83.0\% & - & 5\\
    \textsc{PubChem} & \textsc{GG-LW}          & 0.017 & $\approx \boldsymbol{0}$ & $\approx \boldsymbol{0}$ & \textbf{35.1}/\textbf{37.7} & \textbf{36.1}/\textbf{39.2} & 0.001 & $\approx \boldsymbol{0}$ & 0.252 & 0.048 & 99.5\% & 97.9\% & 8.1\% & 3\\
                     & \textsc{GG-Full}        & 0.005 & $\approx \boldsymbol{0}$ & $\approx \boldsymbol{0}$ & 33.1/37.7& 34.1/39.2 & $\approx \boldsymbol{0}$ &  $\approx \boldsymbol{0}$ & 0.135 & 0.021 & 99.2\% & \textbf{99.5\%} & 15.7\% & 2\\
                     & \textsc{*GG-Redux}       & \textbf{0.001} & $\approx \boldsymbol{0}$ & $\approx \boldsymbol{0}$ & 34.3/37.7 & 35.6/39.2 & $\approx \boldsymbol{0}$ & $\approx \boldsymbol{0}$ & \textbf{0.037} & \textbf{0.009} & \textbf{99.6\%} & 93.7\% & \textbf{51.5\%} & \textbf{1}\\
                
    \midrule
                     & \textsc{GraphRNN}      & 0.103 & 0.301 & 0.043 & 6.3/35.9 & 6.3/37.6 & 0.193 & 0.005 & 0.836 & 0.325 & 86.0\% & 45.0\% & - & 4\\
                     & \textsc{DGMG}          & 0.123 & 0.001 & 0.026 & 11.0/35.9 & 10.3/37.6 & 0.083 & 0.002 & 0.842 & 0.260 & 98.0\% & 98.0\% & - & 4\\
    \textsc{Lung}    & \textsc{GG-LW}         & 0.015 & $\approx \boldsymbol{0}$ & $\approx \boldsymbol{0}$ & \textbf{35.9}/\textbf{36.1} & \textbf{37.7}/\textbf{37.7} & \textbf{0.001} & $\approx \boldsymbol{0}$ & 0.334 & 0.054 & \textbf{99.9\%} & 99.5\% & 9.8\% & 2\\
                     & \textsc{GG-Full}       & 0.007 & $\approx \boldsymbol{0}$ & $\approx \boldsymbol{0}$ & 35.6/35.9 & 37.0/37.6 & 0.002 & $\approx \boldsymbol{0}$ & 0.154 & 0.02 & \textbf{99.9\%} & \textbf{99.9\%} & 3.6\% & 3\\
                     & \textsc{*GG-Redux}      & \textbf{0.001} &  $\approx \boldsymbol{0}$ & $\approx \boldsymbol{0}$ & 34.5/35.9 & 36.1/37.6 & \textbf{0.001} & $\approx \boldsymbol{0}$ & \textbf{0.050} & \textbf{0.009} & 99.6\% & 98.7\% & \textbf{43.6\%} & \textbf{1}\\
                 
    \midrule 
                     & \textsc{GraphRNN}      & 0.512 & 0.153 & 0.026 & 26.9/32.1 & 27.0/33.2 & 0.310 & 0.002 & 0.997 & 0.597 & 93.0\% & 90.0\% & - & 4\\
                     & \textsc{DGMG}          & 0.056 & 0.002 & 0.008 & 34.9/32.1 & 35.1/33.2 & 0.115 & $\approx \boldsymbol{0}$ & 0.967 & 0.239 & 90.0\% & 89.0\% & - & 4\\
  \textsc{Yeast}     & \textsc{GG-LW}         & 0.011 & $\approx \boldsymbol{0}$ & $\approx \boldsymbol{0}$ & \textbf{31.9}/\textbf{32.4} & \textbf{32.6}/\textbf{33.5} & 0.002 & $\approx \boldsymbol{0}$ & 0.208 & 0.040 & 99.2\% & 97.2\% & 15.0\% & 2\\
                     & \textsc{GG-Full}       & 0.008 & $\approx \boldsymbol{0}$ & $\approx \boldsymbol{0}$ & 31.0/32.4 & 31.8/33.5 & \textbf{0.001} & $\approx \boldsymbol{0}$ & 0.133 & 0.030 & \textbf{99.3\%} & \textbf{97.3\%} & 8.1\% & 2\\
                     & \textsc{*GG-Redux}     & \textbf{0.001} & $\approx \boldsymbol{0}$ & $\approx \boldsymbol{0}$ & 30.2/32.4 & 31.2/33.5 & \textbf{0.001} & $\approx \boldsymbol{0}$ & \textbf{0.034} & \textbf{0.009} & 98.2\% & 91.5\%  & \textbf{54.3\%} & \textbf{1}\\
                  
    \midrule
                     & \textsc{GraphRNN}      & 0.243 & 0.198 & 0.016 & 32.4/32.9 & 52.8/64.1 & 0.005 & - & 0.249 & 0.051 & 98.0\% & 99.0\% & - & 4\\
                     & \textsc{DGMG}          & - & - & - & - & - & - & - & - & - & - & - & - & -\\
    \textsc{Enzymes} & \textsc{GG-LW}         & 0.251 & 0.313 & 0.013 & 24.1/32.5 & 38.1/62.2 & 0.086 & $\approx \boldsymbol{0}$ & 0.438 & 0.142 & 99.0\% & 99.3\% & - & 3\\
                     & \textsc{GG-Full}       & 0.210 & 0.260 & \textbf{0.012} & 24.3/32.5 & 39.1/62.2 & 0.097 & $\approx \boldsymbol{0}$ & \textbf{0.412} & 0.135 & \textbf{99.4\%} & \textbf{99.2\%} & - & 2\\
                     & \textsc{*GG-Redux}     & \textbf{0.085} & \textbf{0.205} & 0.022 & \textbf{29.6}/\textbf{32.4} & \textbf{52.4}/\textbf{62.2} & 0.059 & $\approx \boldsymbol{0}$ & 0.422 & \textbf{0.126} & 98.9\% & 99.1\% & - & \textbf{1}\\
    \midrule
                     & \textsc{GraphRNN}      & 0.089 & 0.083 & 0.100 & 36.0/48.6 & 41.8/59.1 & 0.024 & - & 0.032 & 0.020 & 83.0\% & \textbf{95.0}\% & - & 4 \\
                     & \textsc{DGMG}          & - & - & - & - & - & - & - & - & - & - & - & - & -\\
    \textsc{CiteSeer}& \textsc{GG-LW}         & 0.105 & 0.103 & 0.092 & 29.8/43.4 & 33.7/52.2 & 0.020 & - & 0.033 & 0.019 & 95.5\% & 89.9\% & - & 3\\
                     & \textsc{GG-Full}       & 0.077 & 0.087 & 0.071 & 32.2/43.4 & 37.1/52.2 & \textbf{0.011} & - & 0.026 & 0.015 & 96.9\% & 91.0\% & - & 2\\
                     & \textsc{*GG-Redux}     & \textbf{0.035} & \textbf{0.051} & \textbf{0.050} & \textbf{36.7}/\textbf{43.4} & \textbf{41.2}/\textbf{52.2} & 0.012 & - & \textbf{0.017} & \textbf{0.012} & \textbf{97.4\%} & 89.0\% & - & \textbf{1}\\
    \midrule 
                     & \textsc{GraphRNN}      & 0.061 & 0.117 & 0.089 & 48.7/58.4 & 55.8/68.6 & 0.017 & - & 0.013 & 0.012 & 91.0\% & \textbf{98.0}\% & - & 4\\
                     & \textsc{DGMG}          & - & - & - & - & - & - & - & - & - & - & - & - & -\\
    \textsc{Cora}    & \textsc{GG-LW}         & 0.101 & 0.125 & 0.120 &  35.1/52.0 & 39.4/60.7 & 0.023 & - & 0.020 & 0.015 & 97.9\% & 89.2\% & - & 3\\
                     & \textsc{GG-Full}       & 0.056 & 0.072 & 0.084 & 37.3/52.0 & 42.6/60.7 & 0.015 & - & 0.014 & 0.011 
                     & \textbf{98.9\%} & 94.4\% & - & 2\\
                     & \textsc{*GG-Redux}     & \textbf{0.027} & \textbf{0.046} & \textbf{0.055} & \textbf{45.0}/\textbf{52.0} & \textbf{50.1}/\textbf{60.7} & \textbf{0.010} & - & \textbf{0.010} & \textbf{0.010} & 98.5\% & 89.9\% & - & \textbf{1}\\
    \bottomrule
    
    \end{tabular}
        \caption{The results of \textsc{GraphGen-Redux} and the two variants of \textsc{GraphGen} evaluated on the proposed metrics. Each value displayed is the mean of 10 trials (standard deviations not shown), except the Rank column, which ranks the models based on how many times they obtain the best performances in the twelve metrics. Lower values are better. \textsc{GraphRNN} and \textsc{DGMG} results are taken from the \textsc{GraphGen} paper.} \label{tab:results1}
    \end{center}
\end{table*}
\subsection{Evaluation}\label{sec:metrics}
Besides the original \textsc{GraphGen} implementation, we also compare the proposed model with \textsc{GraphRNN} \cite{you2018graphrnn} and the model in \cite{li2018learningdeepgmg} (hereby termed \textsc{DGMG}). The comparison is carried out on a total of twelve metrics. The first five capture the local structure of the generated graphs. In particular, we evaluate (1) node degree distribution, (2) clustering coefficient distribution, (3) orbit count\footnote{Orbits are structural motifs with fixed number of nodes shared between the training and generated samples. In this work, we evaluate 4-orbits.} distribution, (4) average node count, and (5) average edge count. Other three metrics capture whether the model was able to learn to label the graphs correctly. Specifically, we compare the distributions of (6) node labels, (7) edge labels, and (8) joint node labels and degrees. Lastly, to capture global graph similarity, we use the Neighborhood Subgraph Pairwise Distance Kernel (NSPDK) \cite{costa2010nspdk}. The NSPDK measures the alignment between two graphs by matching subgraphs with different radii and distances. These metrics are calculated for each graph in a set of generated graphs, and for each graph in a set of test graphs. The distance between the two obtained samples is then calculated with Maximum Mean Discrepancy (MMD) \cite{gretton2012mmdkernel}.

The evaluation procedure on these metrics is carried out as follows. We first sample 2560 graphs from the model to be evaluated, which we partition in 10 batches of 256 graphs. For each batch, we randomly sample 256 graphs from the test set and evaluate the MMD for every metric on these two samples. Since the test set of the \textsc{Enzymes} dataset is composed of 57 graphs, we use the same procedure but with a sample size of 40, repeating the evaluation 64 times. 

The second set of evaluation metrics concerns quantitative aspects of the generated graphs. These metrics are evaluated on the entire generated sample of 2560 graphs. In particular, we compute:
\begin{itemize}
    \item \textit{novelty}, i.e. the percentage of generated graphs not present in the training dataset;
    \item \textit{uniqueness}, i.e. the percentage of unique graphs (not duplicated) out of the total graphs generated;
    \item \textit{chemical validity}, only on the chemical datasets, namely \textsc{PubChem}, \textsc{Lung}, and \textsc{Yeast}. To compute validity, we transform the generated graphs into SMILES strings \cite{weininger1988smiles}, and we count how many of them are successfully parsed by the SMILES validator of the \texttt{rdkit} Python library \cite{rdkit}.
\end{itemize}

\subsection{Architecture and Hyperparameters}
We evaluate \textsc{GraphGen-Redux} with the same hyperparameters across the different datasets. This choice was made to isolate the extent of our contribution from an excessive tuning \cite{lipton2018troubling}, and to highlight the loose dependence of \textsc{GraphGen-Redux} from the hyperparameters. On this matter, we adopt the hyperparameters setup proposed in the \textsc{GraphGen} paper: all models are trained using the Adam optimizer \cite{kingma2015adam} with initial learning rate of 0.003, annealed by a factor of 0.3 after every 100, 200, 400, and 800 epochs. The RNN layers have been regularized with dropout \cite{srivastava2014dropout}, using a drop rate of 0.2. As regards the \textsc{GraphGen-Redux} architecture, the output size of the linear embedding layer is 64, the hidden state size of each of the four LSTM layers is 128, and all three MLPs have one hidden layer with output size of 128. As regards \textsc{GraphGen}, we trained two different variants. One, which we term \textsc{GraphGen-Full}, uses the same architecture and hyperparameters proposed in the original paper. A second variant, which we term \textsc{GraphGen-LightWeight} (LW), uses the same architecture and hyperparameters as ours. The only difference from the proposed is the number of MLPs necessary to predict the components of the sequence elements (five instead of three). For the \textsc{GraphGen} variants, we use the same number of training epochs as in the related paper. All models are trained with a batch size of 32. 

\subsection{Other Details}
We train \textsc{GraphGen-Redux} and the two \textsc{GraphGen} variants from scratch. This choice has two rationales: first, the evaluation is fairer since both are evaluated on the same machine. Specifically, training and evaluation were carried out on a server with dual Intel Xeon Gold 6132 processor with 14 physical cores each and 384 GB of RAM, using one NVidia Tesla T4 GPU with 16 GB of GPU memory. Second, we could not reproduce the original \textsc{GraphGen} results, since the dataset, the data splits, and the model weights are not provided. In this respect, we make available the datasets, the data splits and the model weights to ensure full reproducibility\footnote{\url{https://github.com/marcopodda/graphgen-redux}}.

\section{Results}
In this section, we provide the results of the experiments, substantiating the claims made in Sec.~\ref{sec:experiments} with empirical evidence.

\subsection{\textsc{GraphGen-Redux} performs better than \textsc{GraphGen}}
Tab.~\ref{tab:results1} shows the results of the experimental evaluation across the considered datasets on the twelve metrics described in Sec.~\ref{sec:metrics}. From the comparison, it is evident how \textsc{GraphGen-Redux} performs consistently better than the two \textsc{GraphGen} variants. This result is summarized in the last column of the table, where the rank of the models is shown. To compute the rank, we simply count how many times a model obtains the best performance across on a given dataset: \textsc{GraphGen-Redux} is highest in rank in all the datasets considered. Our model performs particularly well in the joint node label and degree distribution and the NSPDK metrics. In particular, the latter is designed to assess global graph similarity, and is considered the most important metric \cite{goyal2020graphgen}. We observe that, despite performing worse in all the other metrics, the \textsc{GraphGen-LW} variant seems to capture the right average number of nodes and edges in the chemical datasets. However, and in sharp contrast with the two \textsc{GraphGen} variants, our model is able to generate many more chemically valid graphs (with an increase varying from 3.5 to 14 times). This result confirms that modeling the edge and node dependencies join learning the node and edge labels information jointly, as opposed to independently, is effective at better recognizing patterns of real-world graphs.

\subsection{\textsc{GraphGen-Redux} is lightweight}
\begin{table}[t]
    \scriptsize
    \begin{center}
    \begin{tabular}{l l r c c }
    \toprule
    \parbox[h]{0.8cm}{\centering Dataset}& 
    \parbox[h]{1.3cm}{\centering \textsc{GraphGen}}&  
    \parbox[h]{0.7cm}{\centering Time}& 
    \parbox[h]{0.8cm}{\centering Epochs}& 
    \parbox[h]{0.8cm}{\centering Num.\\Params.}\\
    \midrule
                      & \textsc{LW}     & 9h 30m & 350 & 633k\\
    \textsc{PubChem}      & \textsc{Full}   & 9h 55m & 350 & 2.77M\\
                      & \textsc{*Redux} & \textbf{5h 10m} & \textbf{200} & \textbf{611k}\\
    \midrule
                      & \textsc{LW}     & 3h 50m & 550 & 603k\\
    \textsc{Lung}     & \textsc{Full}   & 4h 05m & 550 & 2.67M\\
                      & \textsc{*Redux} & \textbf{1h 50m} & \textbf{250} & \textbf{580k}\\
    \midrule
                      & \textsc{LW}     & 4h 20m & 250 & 603k\\
    \textsc{Yeast}    & \textsc{Full}   & 4h 00m & 250 & 2.67M\\
                      & \textsc{*Redux} & \textbf{1h 40m} & \textbf{150} & \textbf{580k}\\
    \midrule
                      & \textsc{LW}     & 1h 30m & 4000 & 662k\\
    \textsc{Enzymes}  & \textsc{Full}   & 1h 40m & 4000 & 2.78M\\
                      & \textsc{*Redux} & \textbf{50m} & \textbf{2000} & \textbf{637k}\\
    \midrule
                      & \textsc{LW}     & 1h 50m& 400 & 616k\\
    \textsc{CiteSeer} & \textsc{Full}   & 2h 15m & 400 & 2.72M\\
                      & \textsc{*Redux} & \textbf{1h 25m} & \textbf{200} & \textbf{578k}\\
    \midrule
                      & \textsc{LW}     & 2h 30m & 400 & 621k\\
    \textsc{Cora}     & \textsc{Full}   & 3h 05m & 400 & 2.73M\\
                      & \textsc{*Redux} & \textbf{2h 05m} & \textbf{200} & \textbf{595k}\\
    \bottomrule
    \end{tabular}
    \caption{A comparison between \textsc{GraphGen-Redux} and the two \textsc{GraphGen} variants in terms of training time, number of epochs and number of parameters.} \label{tab:results2}
    \end{center}
\end{table}
Tab.~\ref{tab:results2} compares the three evaluated variants in terms of training times, number of epochs and number of parameters. If we focus on the attention on the number of parameters, we see once again that \textsc{GraphGen-Redux} is the clear winner. In particular, it uses a comparable number of parameters to the \textsc{GraphGen-LW} variant (ranked the lowest in terms of performances), and approximately 78\% less parameters than the second-ranked model (\textsc{GraphGen-Full}). This result certifies that the proposed model is able to obtain comparable or best performances with a lower number of parameters than the original variant.

\subsection{\textsc{GraphGen-Redux} is faster than \textsc{GraphGen}}
Focusing on the Epochs columns of Tab.~\ref{tab:results2}, we see that \textsc{GraphGen-Redux} obtains its performances in approximately half the number of epochs than those required to train the two \textsc{GraphGen} variants. This is reflected on the total time needed to train the model, which scales down proportionally. We also notice that, contrary to our model, the time needed to train the \textsc{GraphGen-LW} variant does not scale down (approximately) linearly with the number of parameters.  

\section{Conclusions}
This paper shows that, with a simple modification to the preprocessing phase, we were able to develop a better version of \textsc{GraphGen} in terms of performances and quality of the generated graphs, which requires a smaller number of parameters and less training epochs to outperform the vanilla variant. Among research directions to explore in the immediate future, we would like to assess whether introducing dependencies in the generation of the timestamps can help produce even more realistically-looking graphs for diverse graph distributions.

\bibliographystyle{ieeetr}
\bibliography{references.bib}   

\end{document}